\documentclass[conference]{IEEEtran}
\usepackage{cite}
\usepackage{amsmath,amssymb,amsfonts}
\usepackage{algorithmic}
\usepackage{graphicx}
\usepackage{textcomp}
\usepackage{xcolor}
\usepackage{subcaption}
\usepackage{booktabs}
\usepackage{siunitx}
\usepackage{etoolbox}
\robustify\bfseries
\usepackage{hyperref}
\usepackage[capitalise]{cleveref}
\usepackage[acronym]{glossaries}
\usepackage{xspace}
\usepackage{blindtext}
\usepackage{multicol}
\usepackage{multirow}

\def\BibTeX{{\rm B\kern-.05em{\sc i\kern-.025em b}\kern-.08em
    T\kern-.1667em\lower.7ex\hbox{E}\kern-.125emX}}

\newacronym{PCG}{PCG}{Procedural Content Generation}
\newacronym{GAN}{GAN}{Generative Adversarial Network}
\newacronym{PCGML}{PCGML}{PCG via Machine Learning}
\newacronym{EA}{EA}{Evolutionary Algorithm}
\newacronym{RL}{RL}{Reinforcement Learning}
\newacronym{MCTS}{MCTS}{Monte Carlo Tree Search}
\newacronym{LSTM}{LSTM}{Long Short-Term Memory}
\newacronym{SAGAN}{SAGAN}{Self-Attention GAN}
\newacronym{WGAN-GP}{WGAN-GP}{Wasserstein GAN with Gradient Penalty}
\newacronym{DCGAN}{DCGAN}{Deep Convolutional GAN}
\newacronym{UMAP}{UMAP}{Universal Manifold Approximation and Projection}
\newacronym{TPKL-Div}{TPKL-Div}{Tile Pattern KL-Divergence}
\newacronym{NLP}{NLP}{Natural Language Processing}
\newacronym{ML}{ML}{Machine Learning}
\newacronym{SMB}{SMB}{Super Mario Bros.}
\newacronym{FID}{FID}{Fr\'echet Inception Distance}

\begin{document}
\newcommand{\red}[1]{\textcolor{red}{#1}}
\newcommand{\methodname}{World-GAN\xspace}
\newcommand{\rulesep}{\unskip\ \vrule\ }
\newcommand{\upsampled}{\!\!\uparrow}

\title{
World-GAN: a Generative Model for Minecraft Worlds
}

\author{\IEEEauthorblockN{Maren Awiszus\IEEEauthorrefmark{1}}
\IEEEauthorblockA{\textit{Institut f\"ur Informationsverarbeitung} \\
\textit{Leibniz University Hannover}\\
Hannover, Germany \\
awiszus@tnt.uni-hannover.de}
\and
\IEEEauthorblockN{Frederik Schubert\IEEEauthorrefmark{1}}
\IEEEauthorblockA{\textit{Institut f\"ur Informationsverarbeitung} \\
\textit{Leibniz University Hannover}\\
Hannover, Germany \\
schubert@tnt.uni-hannover.de}
\and
\IEEEauthorblockN{Bodo Rosenhahn}
\IEEEauthorblockA{\textit{Institut f\"ur Informationsverarbeitung} \\
\textit{Leibniz University Hannover}\\
Hannover, Germany \\
rosenhahn@tnt.uni-hannover.de}
}

\maketitle
\begingroup\renewcommand\thefootnote{\IEEEauthorrefmark{1}}
\footnotetext{Equal contribution}
\endgroup

\begin{abstract}
This work introduces \methodname, the first method to perform data-driven Procedural Content Generation via Machine Learning in \emph{Minecraft} from a single example.
Based on a 3D \gls*{GAN} architecture, we are able to create arbitrarily sized world snippets from a given sample.
We evaluate our approach on creations from the community as well as structures generated with the Minecraft World Generator.
Our method is motivated by the dense representations used in \gls*{NLP} introduced with \emph{word2vec} \cite{mikolovEfficientEstimationWord2013a}. %
The proposed \emph{block2vec} representations make \methodname independent from the number of different blocks, which can vary a lot in Minecraft, and enable the generation of larger levels. %
Finally, we demonstrate that changing this new representation space allows us to change the generated style of an already trained generator.
\methodname enables its users to generate Minecraft worlds based on parts of their creations.
\end{abstract}

\glsresetall


\section{Introduction}
\label{sec:introduction}

\gls*{PCG} has been applied to many different areas and games \cite{snodgrassGeneratingMapsUsing2014,gutierrezGenerativeAdversarialNetwork2020}.
Recently, the progress in Machine Learning has especially spurred research in the field of \gls*{PCGML} \cite{summervilleProceduralContentGeneration2018}.
While there are methods to generate levels for 2D games such as \emph{Super Mario Bros.} \cite{volzEvolvingMarioLevels2018b}, generating levels in 3D is an open area of research.
There has been work on generating levels for the 3D game \emph{Doom} \cite{giacomello2018doom}, but the generation process relies on generating 2D descriptions of the level layout that prohibit complex vertical structures. %

There are 3D games that use grammars and rule-based algorithms for \gls*{PCG} in level generation. %
However, these games do not yet use Machine Learning in their generation process and are thus only extendable by programming new rules.
The most prominent game in this domain is \emph{Minecraft} \cite{minecraft}, which can generate endless worlds in a 3D voxel space.
Minecraft's World Generator is intricately handcrafted to generate vast landscapes, filled with structures like villages or caves with mine shafts.
These landscapes also change with so-called biomes, which define the type of area, from plains to deserts to beaches.
Human generated content also plays an important role in Minecraft, but the structures have to be placed manually in a fixed world and the World Generator cannot learn to reconstruct or generate newly built structures on its own.

\begin{figure}
    \centering
    \includegraphics[width=\linewidth]{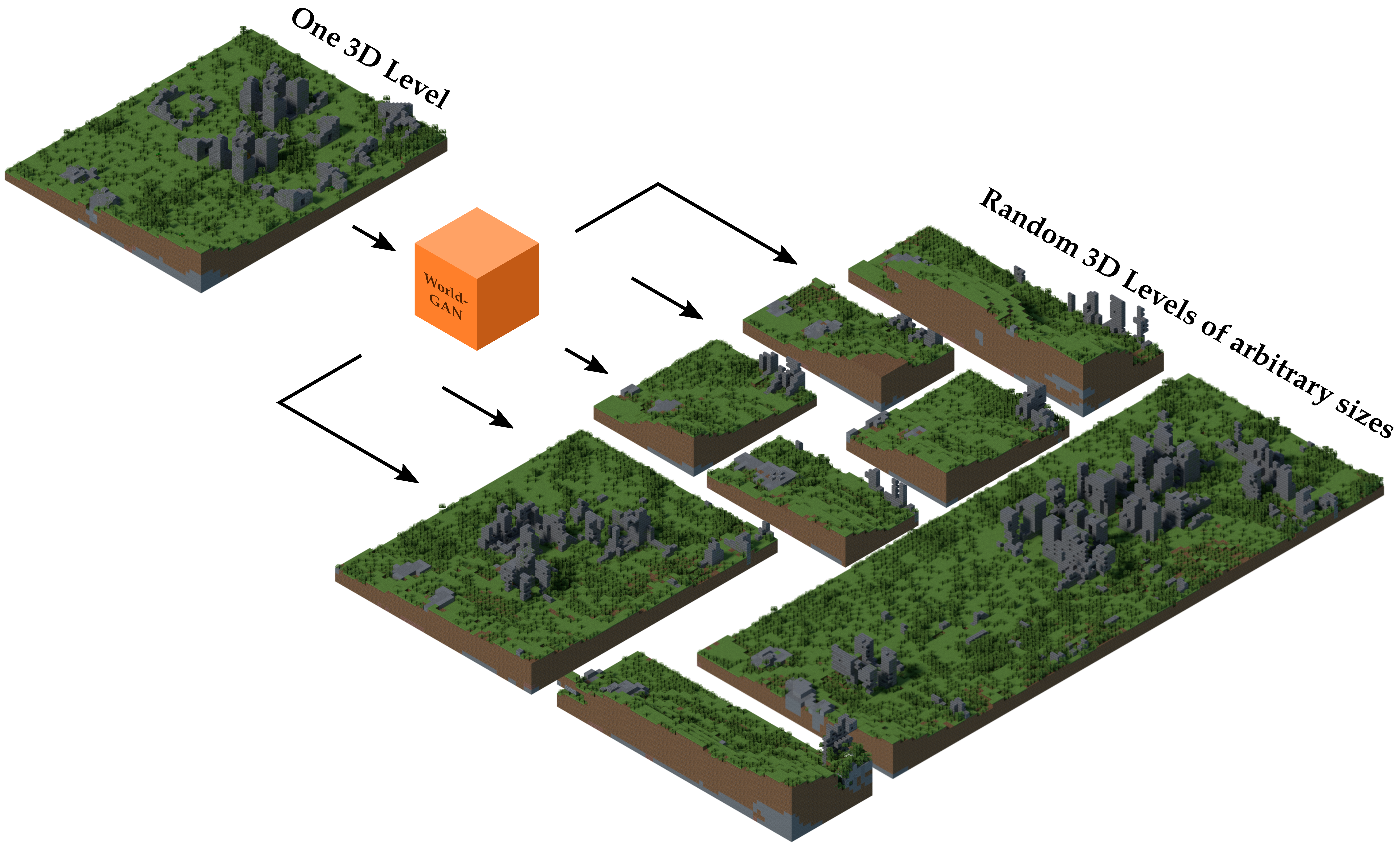}
    \caption{Examples of generated Minecraft world snippets trained on a sample of handcrafted stone ruins in a field of grass.
    The samples can be generated in arbitrary sizes and capture the structures of the ruins and the terrain of the original area.}
    \label{fig:teaser}
\end{figure}

In this paper, we bridge the gap between the rule-based \gls*{PCG} of the Minecraft World Generator and custom structures that were designed by hand, as our proposed method aims to learn and generate structures directly in 3D voxel space (compare \cref{fig:teaser}, all world visualizations are made using Mineways \cite{Mineways} and Blender \cite{Blender}).

In summary, our \textbf{contributions} are:
\begin{itemize}
    \item We introduce a 3D \gls*{GAN} architecture for level generation in Minecraft.
    \item Our proposed dense token\footnote{The smallest building block or tile a level is made of, e.g. sprites or voxels.} representations using \emph{block2vec} enable the processing of larger world snippets.
    \item Editing this representation space allows the application of style changes to generated levels without further training.
    \item Using a current version of Minecraft (1.16) makes our method widely applicable for practitioners.
    \item We enable others to generate their own world snippets by releasing our source code at \url{https://github.com/Mawiszus/World-GAN}.
\end{itemize}

\begin{figure*}
    \centering
    \subcaptionbox{Pipeline \label{fig:pipeline_a}}[.65\linewidth]{
    \includegraphics[width=\linewidth]{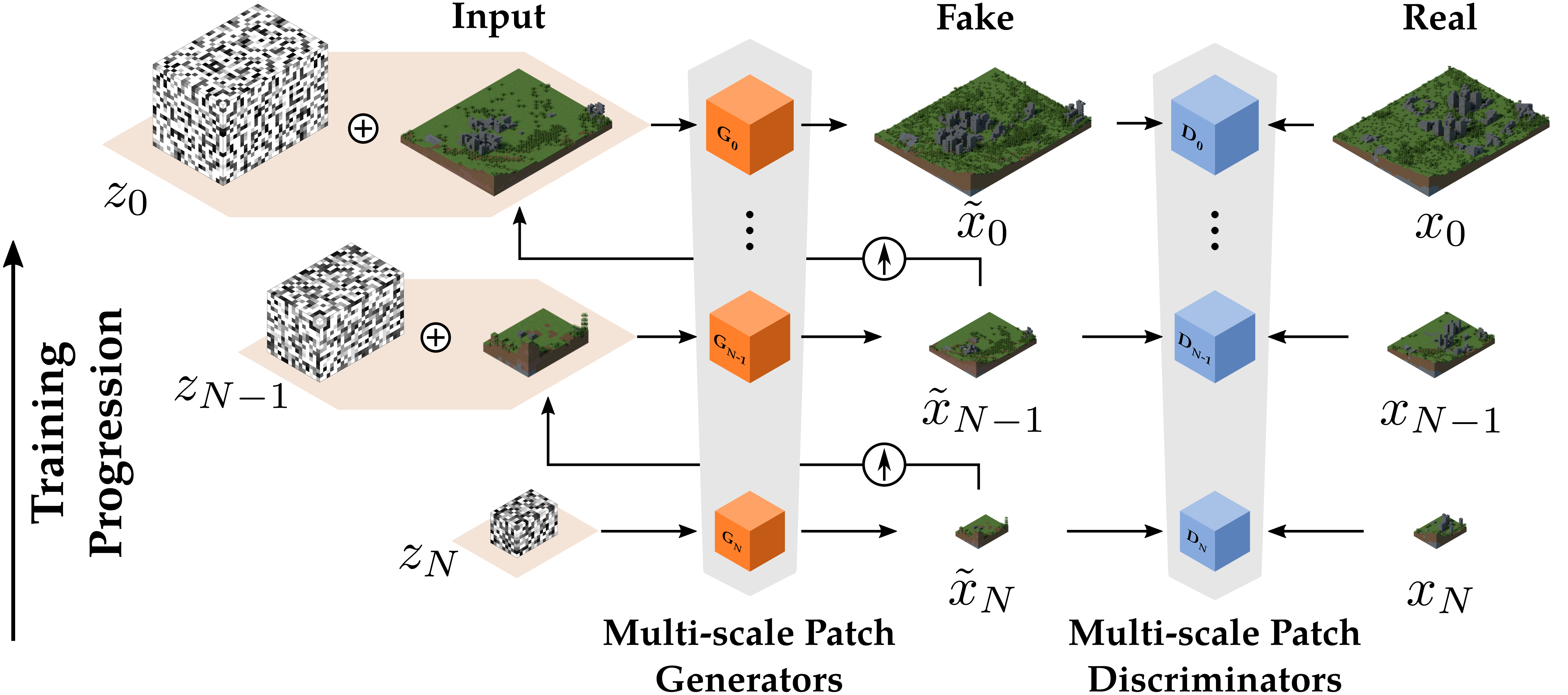}
    \vspace{0.001cm}}
    \rulesep~
    \subcaptionbox{3D Convolution \label{fig:pipeline_b}}[.26\linewidth]{
    \includegraphics[width=\linewidth]{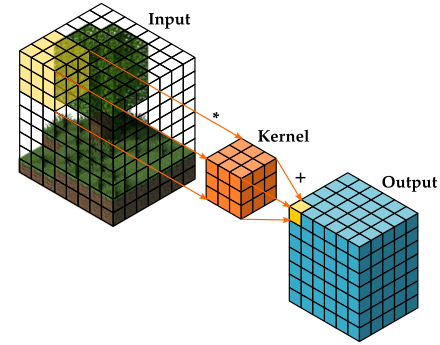}
    \vspace{0.6cm}}
        
    \caption{The \methodname training pipeline. Similar to \cite{awiszus2020toad}, different patch-based generators are trained at different scales to create locally convincing world snippets that the discriminators are trying to distinguish. 
    While the original method uses one-hot encodings for their levels, we use dense representations from block2vec (\cref{subsec:block2vec}) that are mapped to functional levels after training.
    These representations make us independent of the number and types of blocks in a level.
    \methodname uses 3D convolutions to process the 3D structures in the given world snippets.}
    \label{fig:pipeline}
\end{figure*}

\section{Related Work}
\label{sec:related-work}

The field of \gls*{PCGML} has seen many advances in recent years, due to the growing capabilities of Machine Learning algorithms.
Besides classical methods like Markov Random Fields and \glspl*{GAN} \cite{volzCapturingLocalGlobal2020}, its methods have been used for level generation in \emph{Candy Crush Saga} and \emph{\gls*{SMB}} \cite{volzEvolvingMarioLevels2018b}.
A recent approach framed \gls*{PCG} as a \gls*{RL} problem and generated Zelda and Sokoban levels \cite{khalifa2020pcgrl} using a Deep \gls*{RL} agent.
Our method is inspired by our previous work TOAD-GAN \cite{awiszus2020toad}, which extended SinGAN \cite{shahamSinGANLearningGenerative2019} to token-based games by using a hierarchical downsampling operation.

In contrast to these existing studies, we propose a method for 3D level generation in Minecraft that adapts the idea of token embeddings from \gls*{NLP} to overcome memory bottlenecks and manually-defined token hierarchies.
Embeddings of game entities have not been used in \gls*{PCG}, but were posed as a future research direction in \cite{khamenehEntityEmbeddingGame2020}.

A game where \gls*{PCG} plays an essential role is Minecraft \cite{minecraft}.
The complex 3D structures in this game pose a problem for \gls*{PCGML} methods.
The AI Settlement Generation Challenge \cite{salgeAISettlementGeneration2020,salgeGenerativeDesignMinecraft2018,salgeGenerativeDesignMinecraft2019} was recently created to spur research in this direction.
The submitted algorithms are generating villages in a given world and are evaluated using subjective measures (adaptability, functionality, narrative, aesthetics).
The creators of the challenge mention data-driven approaches as a future direction of \gls*{PCG} in Minecraft which was one motivation for our work.
One method to increase the diversity of the generated content was published by Green et al.~\cite{greenOrganicBuildingGeneration2019}, which generates floor plans using a constrained-growth algorithm and Cellular Automata.

Several simplified Minecraft-inspired simulators were proposed \cite{sorosVoxelbuildMinecraftinspiredDomain2017,patrascuArtefactsMinecraftMeets2016} to study the creative space of 3D structures.
Grbic et al.~\cite{grbicEvoCraftNewChallenge2020} introduce the problem of open-ended procedural content generation in Minecraft.
Sudhakaran et al.~\cite{sudhakaranGrowing3DArtefacts2021} use the Neural Cellular Automata architecture to produce a fixed structure in Minecraft given a seed or partial structure.

Yoon et al.~\cite{Yoon2018DesignMF} classify Minecraft villages into several themes (e.g. medieval, futurist, asian) but do not perform \gls*{PCG}.
There have been experiments to generate Minecraft structures based on user-defined content \cite{bluShine2020}, but the results were not satisfying.
Our proposed \methodname is one of the first practical \gls*{PCGML} applications for Minecraft.

\section{Method}
\label{sec:method}

Our method builds upon several existing techniques which are briefly described in this section before we introduce \methodname and our block2vec algorithm.

\subsection{Generative Adversarial Networks}

\methodname is based on the \gls*{GAN} \cite{goodfellow2014generative} architecture.
Given a dataset, these networks are able to generate new samples that are similar to the provided examples.
They are trained by using two adversaries, a generator $G$ and a discriminator $D$.
The generator is fed a random noise vector $z$ and produces an output $\tilde{x}$.
Then, the discriminator is either given a real sample $x$ or the generated one and has to predict whether the sample is from the real dataset or not.
By learning to fool the discriminator, the generator is gradually producing more and more samples that look as if they belong to the training distribution. %
One problem with this architecture is that it requires a lot of data.
Otherwise, it is too easy for the discriminator to distinguish between real and fake samples and the generator is not able to improve its output.

\subsection{SinGAN}

SinGAN \cite{shahamSinGANLearningGenerative2019} enables the generation of images from only one example by using a fully-convolutional generator and discriminator architecture.
Thus, the discriminator only sees one part of the sample and can more easily be fooled by the generator.
Because the field of view in this architecture is limited, long-range correlations can only be modeled by introducing a cascade of generators and discriminators that operate at $N$ different scales.
The samples for each scale are downsampled and the \glspl*{GAN} are trained beginning from the smallest scale $N$
\begin{equation}
    \tilde{x}_N = G_N(z_N).
\end{equation}

This scale defines the global structure of the generated sample, which will be refined in the subsequent scales.
At scales $0 \leq n < N$, the output from the previous scale is upsampled ($\uparrow$) and passed to the scale's generator after disturbing it with a noise map $z_n \sim \mathcal{N}(0, {\sigma_n}^2)$.
The variance of the noise determines the amount of detail that will be added at the current scale by the generator to produce

\begin{equation}
    \tilde{x}_n = \tilde{x}_{n+1}\upsampled + G_n (z_n + \tilde{x}_{n+1}\upsampled).
\end{equation}

At each scale, the discriminator either receives a downsampled real sample $x_n$ or the output of the generator with equal probability.
The gradient of the discrimination loss is then propagated to the discriminator and the generator, which creates the Minimax problem

\begin{equation}
    \min_{G_n} \max_{D_n} \mathcal{L}_{\text{adv}}(G_n, D_n) + \alpha \mathcal{L}_{\text{rec}}(G_n).
\end{equation}

The loss $\mathcal{L}_{\text{adv}}$ is the widely-used \gls*{WGAN-GP} \cite{arjovskyWassersteinGAN2017,gulrajaniImprovedTrainingWasserstein2017a} loss and $\mathcal{L}_{\text{rec}}$ is a reconstruction loss weighted by $\alpha$ which ensures that the \gls*{GAN}'s latent space contains the real sample\footnote{For a more detailed description see \cite{shahamSinGANLearningGenerative2019} and \cite{awiszus2020toad}.}.
After training on one scale has converged, the parameters of the generator and discriminator are copied to the next scale as an initialization.

\subsection{TOAD-GAN}
\label{subsec:toad-gan}

As SinGAN is designed for modeling natural images, its application to token-based games requires some modifications.
TOAD-GAN \cite{awiszus2020toad} introduces several changes to SinGAN's architecture.
Small structures that consist of only a few or a single token would be missing at lower scales due to aliasing by the downsampling operation.
The bilinear downsampling is thus replaced by a special downsampling operation that considers the importance of a token in comparison with its neighbors.
The importance is determined using a hierarchy that is constructed by a heuristic which is motivated by the TF-IDF metric from \gls*{NLP}.
These extensions allow TOAD-GAN to be applied to \gls*{SMB} and several other 2D token-based games.
However, the generation of 3D content requires some changes to the network architecture of TOAD-GAN.
The jump from 2D to 3D means the size of samples will be significantly bigger and since TOAD-GAN is using one-hot encodings of tokens, the required GPU space grows substantially.
This shortcoming is especially apparent in Minecraft, where the high number of tokens can drastically limit the volume that TOAD-GAN is able to generate.
To put this difference into perspective, a one-hot encoded tensor of the original \gls*{SMB} level 1-1 has a shape of $202 \times 16$ with $12$ (out of $28$ possible) different tokens. Taking only the actually present tokens into account, this results in $38,784$ floating point numbers, which take up $0.16$ MB.
The village example by comparison has a shape of $121 \times 136 \times 33$ with $71$ (out of $300+$ possible) different tokens, resulting in $38,556,408$ numbers that require $154.23$ MB to store. If you do not preprocess the data so that only present tokens are taken into account, the difference becomes even more steep.

\subsection{\methodname}
\label{subsec:world_gan}

While the overall architecture of \methodname in \cref{fig:pipeline_a} is similar to TOAD-GAN, the 3D structure of Minecraft levels requires several modifications.
The generator and discriminator now use 3D convolutional filters that can process the $k \times D \times H \times W$ sized slices from the input level.
Here, $k$ is the number of tokens in a level and $D, H$ and $W$ are the depth, height and width of the slice.
\cref{fig:pipeline_b} shows a visualization of the 3D convolution operation.

Another difficulty is the number of tokens in Minecraft and their long-tailed distribution, i.e. some of the tokens only appear a few times in a given sample whereas others (such as \texttt{air}) take up half of the map.
To make \methodname independent of the number of tokens, we turn to a technique from \gls*{NLP}. %

\subsection{block2vec}
\label{subsec:block2vec}

\begin{figure}
    \centering
    \includegraphics[width=\linewidth]{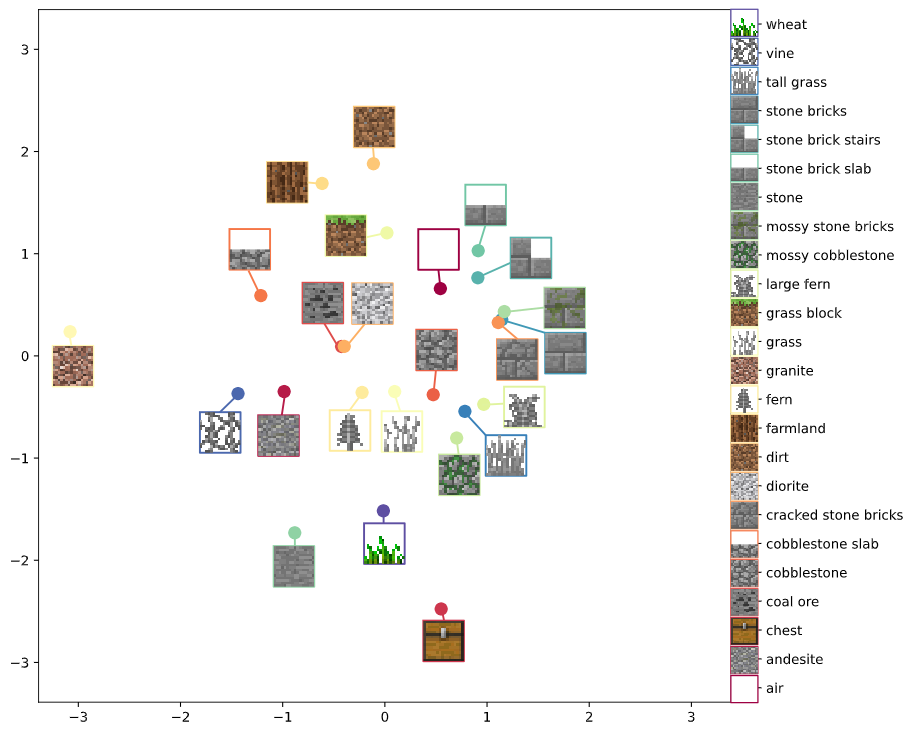}
    \caption{Embeddings learned by block2vec of the ruins structure. The embeddings have 32 dimensions but are transformed to two dimensions for this visualization using the Minimum Distortion Embedding method \cite{agrawal2021minimum}.}
    \label{fig:embeddings_ruins}
\end{figure}

\begin{table}
    \caption{Structure coordinates of our example areas in DREHMAL:PRIM$\Omega$RDIAL \cite{DREHMAL} (visualizations are shown in \cref{fig:samples}).}
    \begin{center}
        \begin{tabular}{l c c c S[table-format=6.0]}
            \toprule
            Structure & x & y & z & {Volume}\\
            \midrule
            desert & [-3219, -3132] & [2628, 2717] & [116, 128] & 92916 \\
            plains & [1082, 1167] & [1110, 1186] & [65, 103] & 245480 \\
            ruins & [1026, 1077] & [1088, 1152] & [63, 73] & 32640 \\
            beach & [606, 695] & [-688, -629] & [39, 64] & 131275 \\
            swamp & [-2753, -2702] & [3242, 3296] & [56, 86] & 82620 \\
            mine shaft & [24987, 25029] & [-799, -754] & [20, 38] & 34020 \\
            village & [25165, 25286] & [-770, -634] & [55, 88] & 543048 \\
            \bottomrule
        \end{tabular}
    \end{center}
    \label{tab:coordinates_structures}
\end{table}

Previous works on \glspl*{GAN} \cite{volzEvolvingMarioLevels2018b,awiszus2020toad} for \gls*{PCGML} use a one-hot encoding of each token in a level.
The downsampling in TOAD-GAN's architecture requires a hierarchy of tokens to enable the generation of small structures at lower scales.
This heuristic, based on term frequencies, has its limitations due to the large number of available tokens in Minecraft which is also constantly expanding in newer versions of the game.
In \gls*{NLP}, this problem was solved by learning a dense fixed-size representation of words \cite{mikolovEfficientEstimationWord2013a}.
These embeddings are constructed by modeling the joint probability of a token and its context.

\begin{figure}
    \centering
    \includegraphics[width=.98\linewidth]{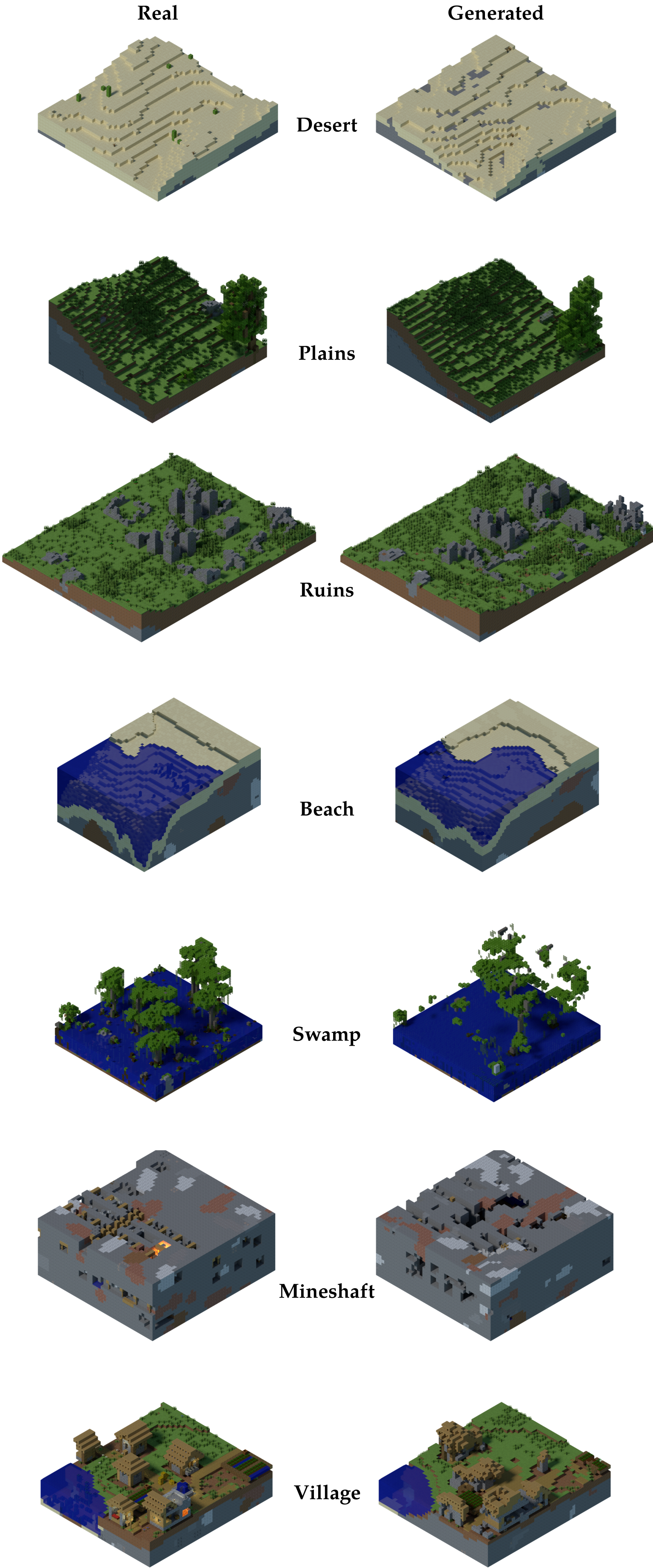}
    \caption{Qualitative samples of our generated levels using block2vec. %
    Using our learned representation space, \methodname is able to create convincing samples of different areas while still including meaningful variations.
    It excels at generating samples in which structures can be freely added and subtracted from, like ruins.
    }
    \label{fig:samples}
\end{figure}

To train these \emph{block2vec} token embeddings, we construct a dataset of blocks and their neighbors from the area of interest, i.e. we create block2vec embeddings for each new area we want to train.
Some tokens such as \texttt{air} occur relatively often, which can lead to sub-optimal representations.
Following \cite{mikolovEfficientEstimationWord2013a}, we mitigate this imbalance by sampling the tokens according to their occurrence probability
\begin{equation}
    P(b_i) = \sqrt{\frac{f(b_i)}{0.001} + 1} \cdot \frac{0.001}{f(b_i)},
\end{equation}
where $f(b_i)$ is the frequency of the token $b_i$ in our given training sample.

We use a skip-gram model with two linear layers, i.e. predicting the context from the target token.
Since our vocabulary is still relatively small in comparison to other \gls*{NLP} tasks, we do not have to employ negative sampling like Mikolov et al.~\cite{mikolovEfficientEstimationWord2013a}.
This algorithm can be seen as a kind of matrix factorization \cite{levyNeuralWordEmbedding} into an $m$-dimensional token representation and a token affinity matrix.
Using a dimensionality reduction technique (MDE) \cite{agrawal2021minimum}, we can visualize our token embeddings (\cref{fig:embeddings_ruins}).

The generators and discriminators in \methodname are given the levels in this representation space, i.e. the generator produces a $m \times D \times H \times W$ tensor that is fed to the discriminator.
After training is complete, the generated levels can be turned into a valid Minecraft level by choosing the token whose embedding is the nearest-neighbor to the generator's output for each voxel.
In contrast to the size calculated in \cref{subsec:toad-gan}, the tensor for the village example now does not depend on the dimensionality of $71$ in the token dimension but on the size of the token embedding. 
We choose $32$ in our experiments, resulting in a tensor using $69.51$ MB instead.
The size of $32$ was empirically chosen, so it can be reduced even further by choosing a smaller representation dimensionality where appropriate.

In addition to reduced memory requirements, block2vec also allows us to omit the definition of a hierarchy.
The token hierarchy was proposed in \cite{awiszus2020toad} to enable the generation of small or rare tokens at lower scales.
Making sure rare tokens are generated is especially important in video game levels, as gameplay relevant secrets or power-ups are usually rare and hidden on purpose.
In our new embedding space, rare tokens are placed close to semantically similar more common tokens.
For example, in the embeddings shown in \cref{fig:embeddings_ruins} the \texttt{stone brick stairs} is a rare token, but its representation is close to one of the most common tokens, \texttt{stone bricks}.
With this, a generator at higher scales can more easily learn to generate the rarer token even if the more common token was generated one scale below.

Finally, choosing a different mapping from internal representations to tokens allows us to change the style of the generated content after training, which we demonstrate in \cref{subsec:repr_editing}.

\section{Experiments}
\label{sec:experiments}

We perform several experiments to evaluate the capabilities of our method, which we will describe in the following sections.
After showing some qualitative samples for a range of different areas in Minecraft, we evaluate several metrics, such as the \gls*{TPKL-Div} and the Levenshtein Distance.
We prove the effectiveness of our block2vec embedding, by comparing it to variants of TOAD-GAN which we extended to 3D.
We call these variants \emph{TOAD-GAN 3D} and \emph{TOAD-GAN 3D*}, for a TOAD-GAN extended to 3D with and without hierarchical\footnote{The hierarchy uses a token frequency based heuristic like in \cite{awiszus2020toad}.} downsampling respectively.
Additionally, we present one version of \methodname that we train on embeddings of the token descriptions from a general purpose \gls*{NLP} model, called BERT \cite{devlinBERTPretrainingDeep2019}. %
Finally, we change the mapping from our representation space to tokens to showcase the possibility of editing \methodname's output after training.

\subsection{Qualitative Examples}
\label{subsec:qual_examples}

Our goal is to generate areas for Minecraft that are similar to a user-defined world snippet but show a reasonable amount of variation.
There is no restriction to what kind of blocks are in the snippet, therefore, any type of area (biomes, buildings, plants) can be used for training \methodname.
To demonstrate the broad applicability of our method we choose a variety of user-defined biomes, like a desert, plains and a beach in our experiments.
We also want to investigate the capability of the method to create simple structures.
For this we select samples with buildings or natural structures like ruins, swamp trees, a mine shaft and a village.
For reproducibility, we extracted all of these samples from the handcrafted world "DREHMAL:PRIM$\Omega$RDIAL" \cite{DREHMAL}, which is available online.
The coordinates in which the areas can be found are shown in \cref{tab:coordinates_structures}.
The village and the mine shaft are drawn from areas of the original Minecraft World Generator.
All other areas are taken directly from hand crafted biomes in \cite{DREHMAL}.

\begin{figure*}
    \centering
    \begin{subfigure}{0.495\textwidth}
        \centering
        \includegraphics[width=.9\linewidth]{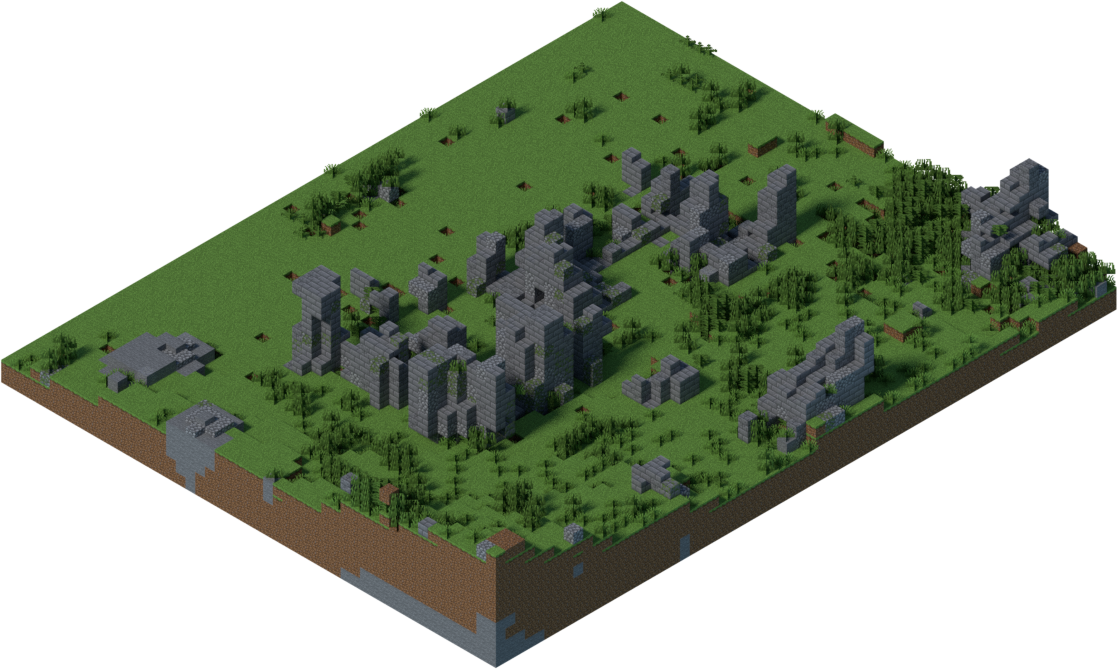}
        \subcaption{TOAD-GAN 3D}
        \label{fig:comparison_hierarchy}
    \end{subfigure}
    \hfill
    \begin{subfigure}{0.495\textwidth}
        \centering
        \includegraphics[width=.9\linewidth]{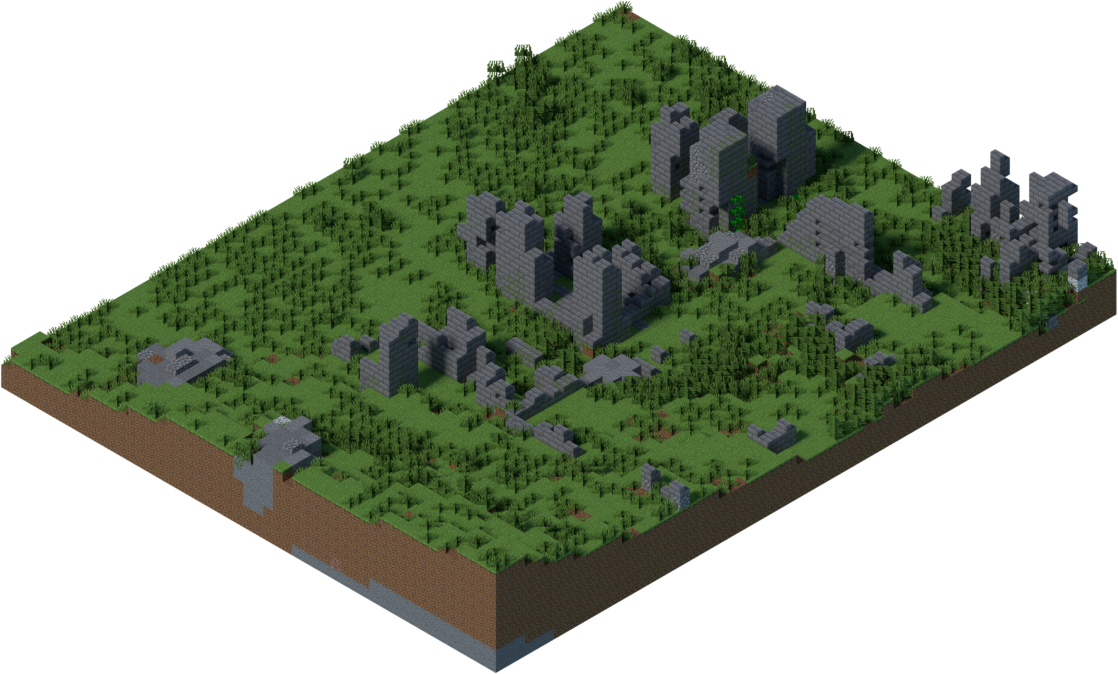}
        \subcaption{\methodname (block2vec)}
        \label{fig:comparison_block2bvec}
    \end{subfigure}
    \caption{Qualitative comparison between content from TOAD-GAN 3D and our block2vec approach.
    (\subref{fig:comparison_hierarchy}) was generated with a simple hierarchy based on the method originally described in \cite{awiszus2020toad}, and (\subref{fig:comparison_block2bvec}) is a sample from the same generator as in \cref{fig:samples}.
    The samples look very similar overall and both methods generate viable structures.
    }
    \label{fig:hierarchy_smaples}
\end{figure*}

\begin{figure}
    \centering
    \includegraphics[width=.99\linewidth]{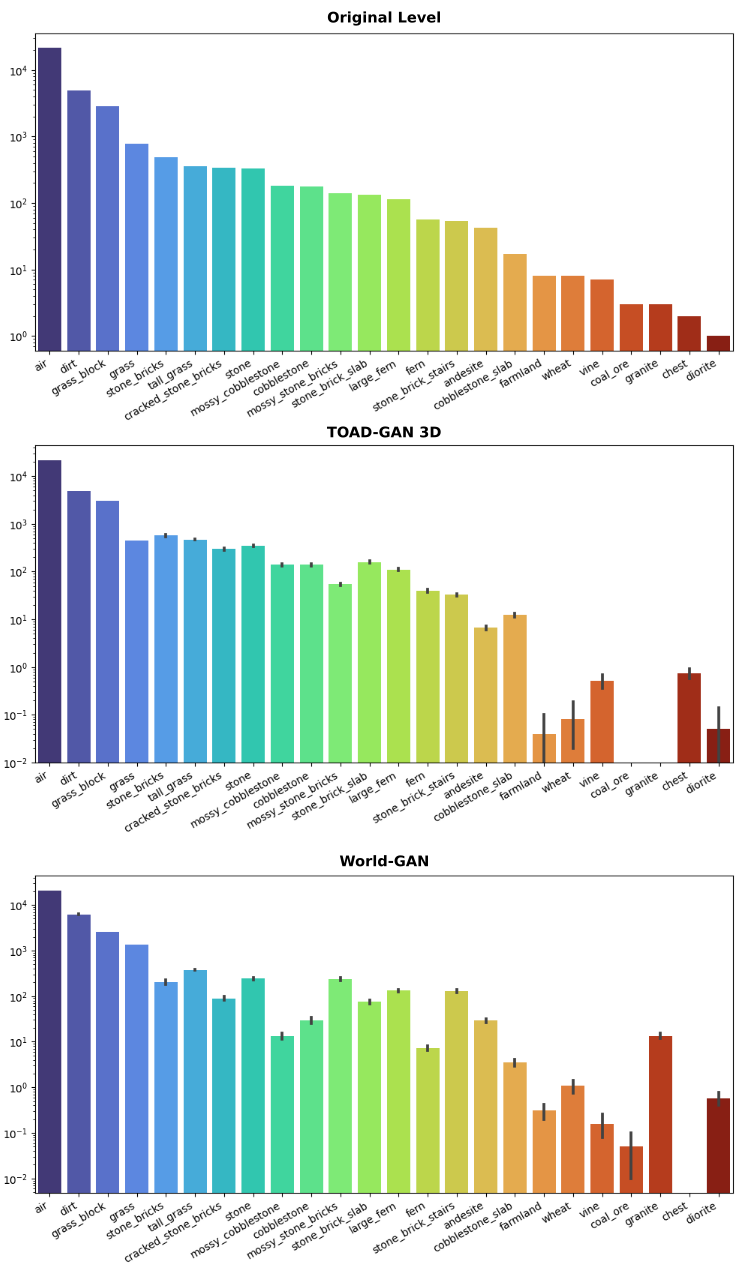}
    \caption{Histograms depicting the block distributions for the ruins example, showing mean and variance over 100 generated samples.
    The y-axis is scaled logarithmically. 
    While there are some differences in the token occurrence counts between our samples and the original, we capture the given distribution reasonably well and are also able to generate most rare tokens.
    }
    \label{fig:block_histogram}
\end{figure}

\cref{fig:samples} shows each of the different areas and one sample generated with our proposed \methodname.
We can see, that \methodname is capable of reproducing a convincing sample of the area while still showing some variability.
Even structures can be generated to a certain degree, such as ruins and trees.
However, as \methodname is not optimized to make coherent, functioning structures, the details of generated houses, like the interior blocks, windows, doors and specific structure are not entirely correct.
Still, the overall structure and style of all areas is captured well by our method.

For comparison, \cref{fig:hierarchy_smaples} shows a sample generated by using a simple hierarchy as in \cite{awiszus2020toad}.
The rank of a block in the hierarchy is defined by the inverse of the sum of its occurrences, i.e. the blocks occurring most often like air and grass blocks are lower in the hierarchy, while blocks that are rare such as stairs and a chest are higher.
We can see that TOAD-GAN 3D also works with its simple hierarchy.
However, it is apparent that rarer blocks are not being generated as much.
Especially in the area where grass and air blocks meet in the upper part of \cref{fig:comparison_hierarchy}, the rarer tall grass block is not generated.

\subsection{Quantitative Evaluation}
\label{subsec:quan_examples}

In this section, we describe three different metrics to evaluate our generated levels: block distribution histograms, the \gls*{TPKL-Div} and the Levenshtein distance.

\subsubsection{Block Distribution}
\label{subsubsec:histograms}

As shown in \cref{subsec:qual_examples}, especially rare tokens are difficult for \gls*{PCGML} methods to model.
By counting the occurrences and visualizing them as histograms we can empirically study whether our method is able to produce rare tokens given the limited number of examples.
\cref{fig:block_histogram} shows how \methodname compares to TOAD-GAN 3D under this metric.
Both methods have varying success generating rare tokens.
Usually, the discriminator will be more likely to label a sample with a rare token as fake.
However, for \methodname with block2vec the rare tokens share some similarities with other more common tokens and are not as easy to detect.
This leads to more of them being generated, as can be seen in the bottom row of \cref{fig:block_histogram}.
Only the \texttt{chest} token is missing from \methodname's output.
As its embedding in \cref{fig:embeddings_ruins} is also further away from the other tokens, we hypothesise that it might be too easy for the discriminator to detect compared to the other rare tokens.
Developing different embeddings that generalise better between common and rare tokens is a direction for future work.

\subsubsection{Tile Pattern KL-Divergence}
\label{subsubsec:TPKL-Div}

\begin{table}%
    \caption{Average Tile-Pattern KL-Divergence between the real structure and 20 generated levels. A lower \gls*{TPKL-Div} implies that the patterns of the original level are matched better.}
    \begin{center}
        \begin{tabular}{l S[table-format=2.2] S[table-format=2.2] S[table-format=2.2]}
            \toprule
            & {\methodname} & {TOAD-GAN 3D} & {TOAD-GAN 3D*} \\
            \midrule
            desert & \bfseries 16.283 & 18.183 & 18.560 \\
            plains & 23.050 & \bfseries 22.791 & 22.844 \\
            ruins & \bfseries 16.196 & 16.354 & 16.507 \\
            beach & 16.764 & \bfseries 15.804 & 16.221 \\
            swamp & 20.320 & \bfseries 18.406 & 19.952 \\
            mine shaft & 14.673 & 14.644 & \bfseries 14.620 \\
            village & 21.728 & 21.573 & \bfseries 21.353 \\
            \midrule
            Average & 18.430 & \bfseries 18.251 & 18.5796 \\
            \bottomrule
        \end{tabular}
    \end{center}
    \label{tab:tpkldiv_structures}
\end{table}

Next, we evaluate how well the patterns in our generated content match the original sample based on the metrics used in \cite{awiszus2020toad}.
The \gls*{TPKL-Div} \cite{lucasTilePatternKLDivergence2019} is the Kullback-Leibler Divergence of token patterns of size $n$ that occur in a level.
We apply it to Minecraft by considering all $n \times n \times n$ patterns in our generated content
\begin{equation}
    D_{KL}(P||Q) = \sum_{s \in \mathbb{N}^{n\times n \times n}} P(s) \log \frac{P(s)}{Q(s)},
\end{equation} where $P(s)$ is the frequency of the pattern $s$ in the original level and $Q(s)$ is its frequency in the generated level.
We choose patterns of size $5 \times 5 \times 5$ and $10 \times 10 \times 10$ and average their resulting \glspl*{TPKL-Div}.
In preliminary experiments we found that using 4 scales of sizes 1.0, 0.75, 0.5 and 0.25 in our generation process leads to the best results\footnote{We evaluated \methodname with scales (1.0, 0.5), (1.0, 0.75, 0.5) and (1.0, 0.75, 0.5, 0.25). The corresponding results are published with our source code.}.
We use this configuration for all qualitative samples that are presented in the paper.
The results in \cref{tab:tpkldiv_structures} show that \methodname with block2vec is able to match the training patterns as well as the other variants while requiring less memory (compare \cref{subsec:world_gan}).

\subsubsection{Levenshtein Distances}
\label{subsubsec:levenshtein}

Finding an objective measure for the uniqueness of procedurally generated content is no easy task.
The Levenshtein distance \cite{levenshtein1966} is an established metric coming from information theory to measure the similarity of two discrete strings.
It is defined as the minimum number of insertions, deletions and substitutions that are needed to transform one string into the other.
The distance is bounded from above by the length of the longer string and is equal to zero iff. the two strings are equal.
We can interpret slices of our generated levels as strings by concatenating the tokens at each position and assigning them a number.
This allows us to compute the Levenshtein distance between all generated samples of an area, which lets us quantify their variability.
The results are shown in \cref{tab:levenshtein_structures}.
\begin{table}
    \centering
    \caption{Average Levenshtein distance between the generated levels. A larger distance implies a larger variability in the generated output.}
    \begin{tabular}{l S[table-format=5.2] S[table-format=5.2] S[table-format=5.2]}
    \toprule
    & {\methodname} & {TOAD-GAN 3D} & {TOAD-GAN 3D*} \\
    \midrule
    desert & \bfseries 3251.370 & 1342.240 & 1305.190 \\
    plains & \bfseries 5314.425 & 3895.960 & 4372.405 \\
    ruins & 5073.485 & \bfseries 5808.965 & 5615.670 \\
    beach & \bfseries 15623.565 & 12846.660 & 11743.380 \\
    swamp & 7900.030 & \bfseries 10515.400 & 7292.570 \\
    mine shaft & \bfseries 9691.685 & 6138.710 & 7764.370 \\
    village & 5721.880 & 6452.610 & \bfseries 6679.685 \\
    \midrule 
    Average & \bfseries 7510.923 & 6714,364 & 6396.183 \\
    \bottomrule
    \end{tabular}
    \label{tab:levenshtein_structures}
\end{table}
We find that content generated using block2vec has a higher variability on average.
One explanation could be the generation of rare tokens, which are more frequent in \methodname's output.
This supports our findings in \cref{subsubsec:histograms}.

\begin{figure}
    \centering
    \includegraphics[width=.88\linewidth]{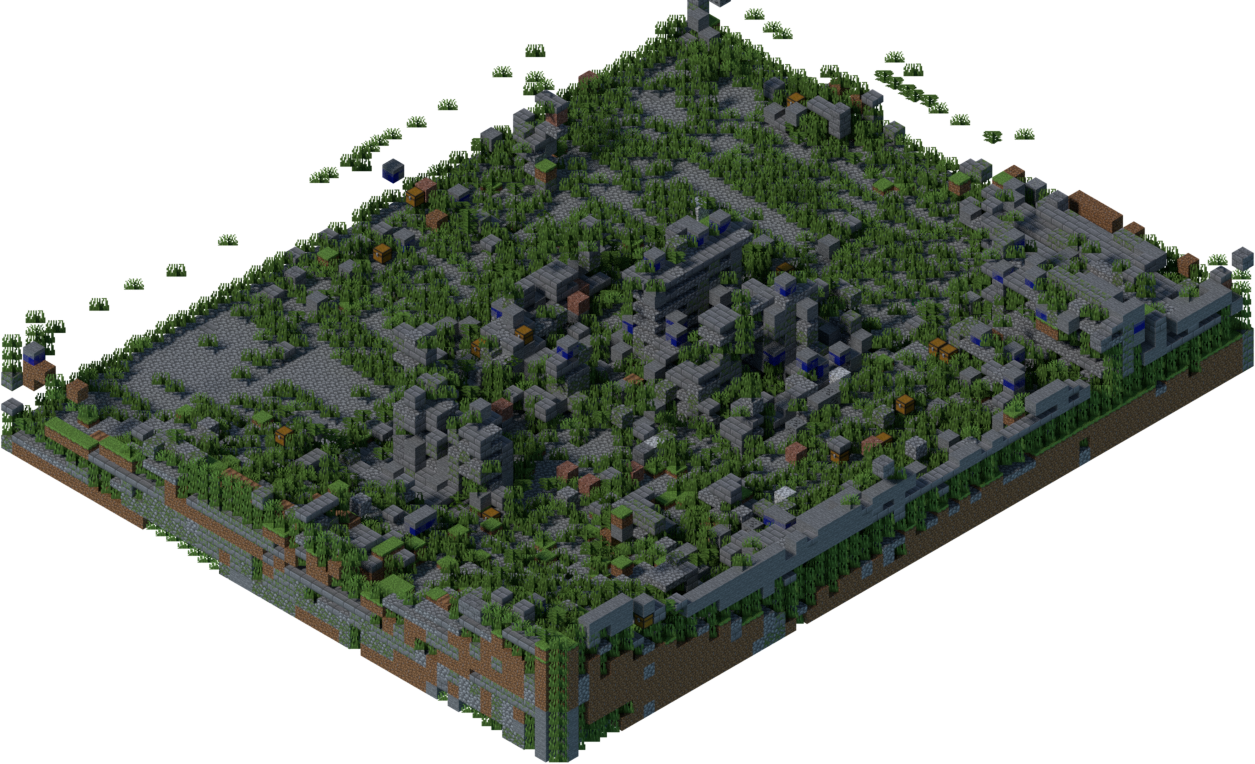}
    \caption{Sample generated with a \methodname trained using BERT embeddings of the token descriptions. 
    Some artifacts such as floating \texttt{tall grass} are produced but the overall structure of the map is visible.
    This indicates that meaningful level generation within a learned word space is possible.
    }
    \label{fig:bert_ruins}
\end{figure}

\subsection{Alternative Token Embeddings}
\label{subsec:alt_token_emb}

We evaluate the impact of the token embeddings on the generated output by comparing our block2vec approach to the canonical BERT \cite{devlinBERTPretrainingDeep2019} embeddings of the token descriptions.
BERT is a widely used \gls*{NLP} model trained on many sample sentences of the English language and can therefore provide embeddings for any English sentence.
For this experiment, we feed the token descriptions (e.g. \texttt{mossy stone bricks}) to a pretrained BERT model \cite{wolf-etal-2020-transformers} and use the final layer's output as our token embedding.
\cref{fig:bert_ruins} shows a qualitative sample when training with these 768 dimensional embeddings.
It is apparent that the patterns are not as closely modeled as with the block2vec embeddings. %
This could be attributed to the high dimensionality of the embeddings. %
However, despite not being trained on Minecraft, the general structure of stone ruins with grass around them is still visible.
Since the embeddings are created using only their textual description, this experiment indicates a future research direction of grounding \methodname in natural language.
This is especially interesting regarding the recently introduced Chronicle Challenge \cite{salgeGenerativeDesignMinecraft2019}.

\subsection{Representation Editing}
\label{subsec:repr_editing}
Similar to TOAD-GAN, levels generated with \methodname can be edited during generation.
The editing capabilities discussed in \cite{awiszus2020toad} are all applicable to \methodname as well.
In this section, however, we want to highlight another possibility of editing the generated levels.
\begin{figure}
     \centering
         \includegraphics[width=.88\linewidth]{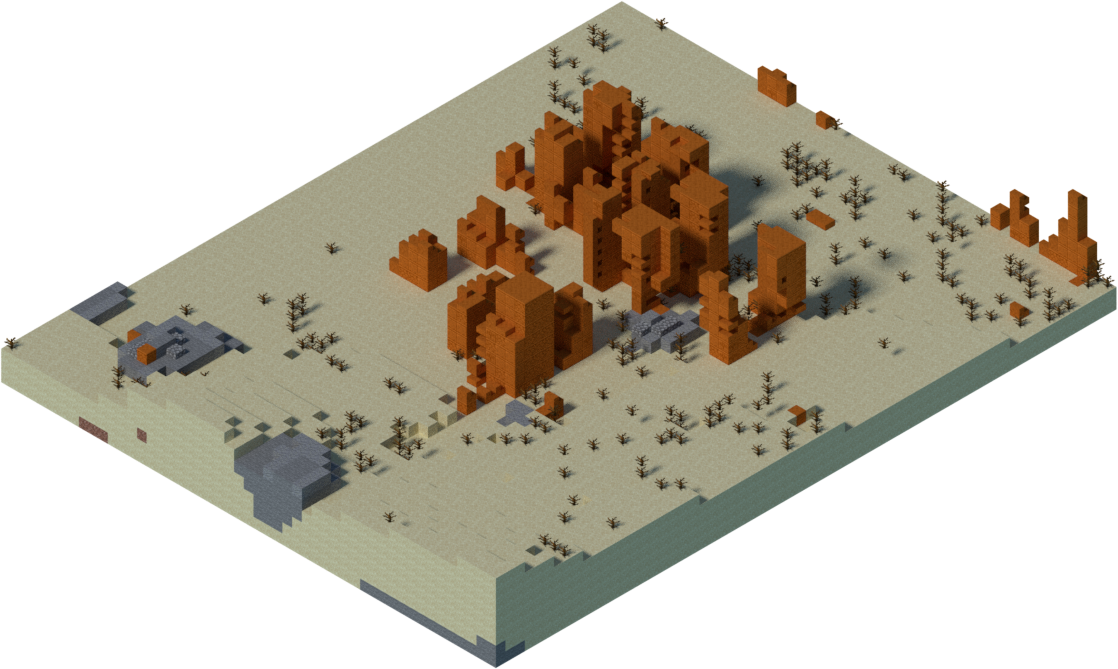}
    \caption{An example of a level with edited representations. 
    With block2vec, we can change the style of the generated structure (ruins) from one style (plains) to another (desert).
    }
    \label{fig:repr_editing}
\end{figure}
As described in \cref{subsec:block2vec}, \methodname is trained on a specific representation space. 
Since the generated samples also belong to this representation space, the transformation from that space back to the original tokens is crucial to the style of the generated level.
By changing this interpretation of the latent space, we can change the style of the generated tokens without retraining the generator.
This results in levels that have the same structure as the original training sample, but can use wildly different tokens.
\cref{fig:repr_editing} shows an example of such a transfer. 
The \methodname used in this example is the same generator used for the ruins examples in previous sections.
For each token vector in the representation space we change the token it is mapped to, in order to fit the new style. 
In this example, we changed the blocks representing the ground, i.e. grass and dirt, to be interpreted as sand and the blocks representing the ruins, i.e. stone bricks, to variations of red sandstone.
This method allows a designer to use the generator of one basic structure for many different styles.

\section{Conclusion and Future Work}
\label{sec:conclusion_and_future_work}

We introduce \methodname, a method that enables data-driven \gls*{PCGML} in Minecraft.
It is inspired by the TOAD-GAN \cite{awiszus2020toad} architecture, but is specifically extended to 3D by incorporating 3D convolutions and the application of dense token embeddings, which we call block2vec.
We evaluated its generated content with respect to its pattern similarity to the original input, its variability and by how well it handles rare tokens in the input.
Finally, we present an easy way to change the representations after training, which enables us to edit the style of a given level.
Especially with the Minecraft Settlement Generation Challenge \cite{salgeGenerativeDesignMinecraft2018} in mind, a few research directions open up for future work.
With the current method, semantic correctness of structures is not enforced, which can result in for example nonsensical houses.
We want to investigate improving our method in order to better generate semantic structures by incorporating semantic rules into the generation process.
Our method opens up several directions for \gls*{PCGML} in Minecraft, as we will publish our source code and are able to handle the most current Minecraft version.

\section{Acknowledgment}
This work has been supported by the Federal Ministry of Education and Research (BMBF), Germany, under the project LeibnizKILabor (grant no. 01DD20003), the Federal Ministry for Economic Affairs and Energy under the Wipano programme "NaturalAI" (03THW05K06), the Center for Digital Innovations (ZDIN) and the Deutsche Forschungsgemeinschaft (DFG) under Germany’s Excellence Strategy within the Cluster of Excellence PhoenixD (EXC 2122).

\bibliographystyle{IEEEtran}
\bibliography{main}

\end{document}